\documentclass[sigconf,nonacm,balance=false]{acmart}
\AtBeginDocument{%
  }

\setcopyright{none}
\usepackage{booktabs}
\usepackage{pgf-pie}
\usepackage{float}

\definecolor{WZBlue}{HTML}{80B1ED}
\definecolor{WZGreen}{HTML}{62C4A5}
\definecolor{WZSand}{HTML}{F3BE66}
\definecolor{WZPurple}{HTML}{B397DE}
\definecolor{WZRose}{HTML}{EC92A7}
\definecolor{WZSlate}{HTML}{6BC8DB}

\newcommand{\wzlegend}[2]{%
  \textcolor{#1}{\rule{0.8em}{0.8em}}\enspace #2%
}

\begin{document}

\title{Trajectory Prediction for Autonomous Driving in Road Construction Zones Using Human-in-the-Loop Driving Data}

\author{Shuo Zhang}
\authornote{Shuo Zhang and Aman Sunesh contributed equally to this work.}
\email{sz4580@nyu.edu}
\affiliation{%
  \institution{New York University}
  \city{New York City}
  \state{NY}
  \country{USA}
}

\author{Aman Sunesh}
\authornotemark[1]
\email{s18181@nyu.edu}
\affiliation{%
  \institution{New York University}
  \city{New York City}
  \state{NY}
  \country{USA}
}

\author{Semiha Ergan}
\email{semiha@nyu.edu}
\affiliation{%
  \institution{New York University}
  \city{New York}
  \country{USA}
}

\author{Kaan Ozbay}
\email{kaan.ozbay@nyu.edu}
\affiliation{%
  \institution{New York University}
  \city{New York City}
  \state{NY}
  \country{USA}
}

\begin{abstract}
Ego-trajectory prediction in urban roadway work zones
remains an underexplored topic in autonomous-driving research, with limited dedicated data and systematic evaluation.
Roadway work zones involve temporary changes in the lane
layouts and connectivity of the roads, complicating
interactions with surrounding vehicles and workers.
These changes make future vehicle trajectories difficult
to predict. To address these challenges, we collect
\textsc{WZDrive-Data}, a multimodal dataset comprising
1,183 work-zone driving segments from 33 drivers,
using a human-in-the-loop driving simulator with
prerecorded worker animations and VR-controlled workers.
Driver control and gaze signals are recorded from human
participants, while vehicle trajectories and scene
information are obtained from the simulator. Our proposed multimodal ego-trajectory predictor, \textsc{WZDrive}, achieves a mean minADE$_6$ of 1.09~m
and minFDE$_6$ of 2.12~m. Compared with the best-performing state-of-the-art baseline for each metric in our benchmark,
WZDrive reduces minADE$_6$ by 7.18\% and minFDE$_6$ by 13.18\%. This research addresses a gap in autonomous-driving research on urban work zones by combining human-in-the-loop
data collection with multimodal learning from human
driving signals and in the context of work-zones. A project website with code and example data is available at
\url{https://shuo9898-zs.github.io/wzdrive/}.
\end{abstract}


\maketitle

\section{Introduction}

Roadway work zones are essential for maintaining and upgrading
transportation and utility infrastructure, but they temporarily change the
conditions under which vehicles operate. Maintenance activities can narrow
or redirect travel lanes and require drivers to respond to surrounding
vehicles, temporary traffic-control devices, and roadway workers in
constrained spaces. These temporary changes can create safety-critical
conditions for drivers and workers. In the United States, an estimated
101,000 work-zone crashes occurred in 2023, resulting in roughly 39,000
injuries and about 900 fatalities \cite{workzone_data2026,nhtsa_crss}. As aging infrastructure and continuing utility work sustain the need for maintenance and rehabilitation \cite{ullman2014workzoneits}, roadway work zones will remain a recurring part of road travel and an important setting for studying how vehicles adapt to temporary road configurations.

Ego-trajectory prediction estimates a vehicle's near-future
motion to support autonomous driving (AD) and advanced
driver-assistance systems (ADAS).
In work zones, these predictions can help AD systems
assess where the vehicle is likely to travel and plan
safer paths around workers and nearby traffic.
For ADAS, they can support hazard warnings and corrective
control when a driver's actions may lead to unsafe motion.
However, temporary lane changes and interactions with
vehicles and workers make prediction challenging.
Limited road space in urban areas can restrict standard
traffic-control layouts, while short-term work zones
may provide less warning and protection, leaving drivers
less time to respond.

Dedicated work-zone datasets have advanced scene
understanding and navigation.
EPOSH focuses on perspective-view and bird's-eye-view
segmentation \cite{dwivedi2021bird}, while ROADWork
provides rich visual annotations and reconstructed
driving paths \cite{ghosh2025roadwork}.
However, ROADWork's pathway-prediction task uses
spatially sampled waypoints derived from video-based
reconstruction rather than directly recorded vehicle
states at fixed time intervals.
General driving datasets, including nuScenes,
Argoverse, and the Waymo Open Motion Dataset,
provide temporal trajectories and established forecasting
benchmarks \cite{caesar2020nuscenes,chang2019argoverse,ettinger2021waymo}. Yet their collection and evaluation protocols are not
specifically organized around safety-critical work-zone
conditions. This leaves a need for dedicated trajectory data and systematic evaluation of vehicle motion under these
conditions.

Human signals offer a potential means of addressing
prediction challenges that trajectory history alone
cannot resolve.
Prior work has developed immersive platforms for
collecting human driving signals \cite{silvera2022dreyevr}
and explored EEG-based feedback and attention-based
guidance for ADAS and autonomous driving
\cite{zhuang2026neurocognitive,zhuang2025aegis,biswas2024gaze}.
These studies motivate investigating human signals
as complementary information for predicting driving
responses in complex work-zone conditions.
We therefore propose a multimodal approach that combines
historical driver-control and attention signals with
vehicle trajectories, surrounding agents, lane geometry,
and temporary work-zone context. To address ego-trajectory prediction in urban roadway
work zones, we recruited 33 participants to collect
driving data across six scenarios using a CARLA--SUMO-based
driving simulator with VR-controlled workers and
prerecorded worker animations
\cite{dosovitskiy2017carla,lopez2018sumo}.

Our contributions includes:
\begin{itemize}
   \item \textbf{Dataset:} We collect \textsc{WZDrive-Data},
    comprising 1,183 work-zone driving segments with
    synchronized vehicle trajectories, driver signals,
    and scene context across six scenario conditions.

%     \item \textbf{Benchmark:} We establish an ego-trajectory
% prediction benchmark for roadway work zones, comparing
% rule-based baselines, standard sequence models,
% and state-of-the-art trajectory predictors.

\item \textbf{Benchmark:} We benchmark rule-based,
sequence, and state-of-the-art models for ego-trajectory
prediction in roadway work zones.

    \item \textbf{Model:} We propose \textsc{WZDrive},
    which combines multimodal encoding and stage-wise
    refinement to improve ego-trajectory prediction
    over the evaluated baselines.

\end{itemize}

%%%%%%%%%%%%%%%%%%%%%%%%%%%%%%%%%%%%%
\section{Related Work}
\label{sec:related_work}

\subsection{Trajectory Prediction}

Trajectory prediction supports autonomous driving and navigation among
human road users by estimating future motion from observed histories
and scene context. Trajectron++ models multimodal agent behavior
\cite{salzmann2020trajectronpp}, while Social-STGCNN captures pedestrian
interactions through spatiotemporal graphs
\cite{mohamed2020socialstgcnn}. Map-aware approaches incorporate lane
structure and agent interactions through lane graphs, graph traversals,
and hierarchical representations
\cite{liang2020lanegcn,deo2022pgp,zhou2022hivt}.

Attention-based models jointly represent agents, time, and heterogeneous
scene inputs \cite{yuan2021agentformer,nayakanti2023wayformer}.
MTR, MTR++, and QCNet use query-based prediction and refinement
\cite{shi2022mtr,shi2024mtrpp,zhou2023qcnet}, while Forecast-MAE
pretrains representations through masked trajectory and lane
reconstruction \cite{cheng2023forecastmae}.
LAFormer combines lane-aware constraints with two-stage refinement
\cite{liu2024laformer}, and DeMo separates directional intentions
from dynamic states \cite{zhang2024demo}.
For ego-trajectory prediction, RouteFormer incorporates trajectory
history, egocentric imagery, and driver field of view
\cite{akbiyik2025routeformer}. These methods primarily address
general driving environments rather than the joint use of temporary
work-zone geometry and human-driver control signals.

% \subsection{Driving and Roadway-Work-Zone Datasets}

% NuScenes, Argoverse, and the Waymo Open Motion Dataset provide
% trajectories and map context across diverse real-world traffic
% conditions \cite{caesar2020nuscenes,chang2019argoverse,ettinger2021waymo}.
% However, they are not dedicated benchmarks for work-zone trajectory
% prediction. Driver-centered datasets add human-related information:
% DR(eye)VE includes driving video, vehicle states, and gaze
% \cite{alletto2016dreyeve}, while RouteFormer's GEM dataset synchronizes
% egocentric video, corrected ego locations, and gaze
% \cite{akbiyik2025routeformer}. Neither specifically targets work zones.

% Work-zone datasets primarily support perception and scene understanding.
% Early research identified work-zone boundaries and traffic changes
% from signs \cite{seo2015workzones}. EPOSH provides egocentric video
% with perspective and bird's-eye-view semantic annotations
% \cite{dwivedi2021eposh}. ROADWork includes diverse work-zone imagery,
% object and sign annotations, scene descriptions, navigational goals,
% and estimated passable paths \cite{ghosh2025roadwork}.
% Dash2Sim reconstructs metric, geo-referenced 4D driving logs from
% dashcam videos and introduces ROADWork4D-CL for closed-loop planning
% evaluation \cite{ghosh2026dash2sim}. These resources do not provide
% the combination of controlled human-driving trajectories, synchronized
% driver controls, and explicit work-zone geometry considered here.

\subsection{Driving and Roadway-Work-Zone Datasets}

General driving datasets, including nuScenes, Argoverse,
and the Waymo Open Motion Dataset, provide trajectories
and map context for motion forecasting
\cite{caesar2020nuscenes,chang2019argoverse,ettinger2021waymo}.
Driver-centered datasets additionally capture human
behavior and attention.
DR(eye)VE includes driving video, vehicle states,
and gaze \cite{alletto2016dreyeve}, while GEM,
introduced with RouteFormer, synchronizes egocentric
video, corrected ego locations, and gaze
\cite{akbiyik2025routeformer}.
Zhuang et al. collect EEG, gaze, and driver-control
data in VR driving simulations to support
human-informed autonomous vehicle control
\cite{zhuang2026neurocognitive}.
These resources support complementary aspects of
driving research but do not specifically target
human driving responses in roadway work zones.

Work-zone research spans perception, scene understanding,
and navigation.
Early work identified work-zone boundaries and traffic
changes from signs \cite{seo2015workzones}.
EPOSH provides perspective-view and bird's-eye-view
semantic annotations \cite{dwivedi2021bird}.
ROADWork adds object and sign annotations, scene
descriptions, and reconstructed driving paths
\cite{ghosh2025roadwork}.
More recently, Dash2Sim reconstructs metric,
geo-referenced 4D driving logs from dashcam videos
and introduces ROADWork4D-CL for closed-loop planning
evaluation \cite{ghosh2026dash2sim}.
Our dataset complements these resources by pairing
controlled human-driving trajectories with synchronized
driver signals and explicit work-zone geometry
for ego-trajectory prediction.

\subsection{Human-in-the-Loop Simulation for Road-User Interaction}

Immersive simulation enables controlled collection of interactions
that would be difficult or unsafe to reproduce in real traffic.
Weng et al. evaluated VR-based collection of pedestrian trajectories
and body poses in safety-critical interactions
\cite{weng2023evaluating}. JaywalkerVR and its CARLA-VR dataset
further support controlled studies of pedestrian behavior
\cite{mukoya2024jaywalkervr}. Related VR frameworks investigate
human-driver--pedestrian interactions, trajectory and crossing
prediction, and coupled intentions and actions
\cite{crosato2024vrframework,dang2024coupling}.
Dang et al. additionally model bounded rationality in game-theoretic
vehicle--pedestrian decision making \cite{dang2025dynamic}.

% Driver-centered platforms provide complementary capabilities.
% DReyeVR integrates immersive driving, eye tracking, driver controls,
% and experiment replay \cite{silvera2022dreyevr}.
% VR-WISE combines driving and work-zone simulation with gaze and
% physiological sensing to study driver awareness
% \cite{zhang2025vrwise}. DualTake examines the transfer of
% takeover-behavior models from simulated car driving to micromobility
% \cite{zheng2024dualtake}. These studies focus on pedestrian behavior,
% driver attention, or takeover prediction rather than work-zone
% ego-trajectory benchmarks.

Driver-centered platforms demonstrate the feasibility
of immersive simulation for multimodal human-driving
research.
DReyeVR integrates CARLA with VR driving, eye tracking,
driver controls, and experiment replay
\cite{silvera2022dreyevr}.
DualTake combines immersive simulation with gaze,
physiological, and vehicle signals to predict takeover
behavior across cars and micromobility
\cite{zheng2024dualtake}.
For roadway work zones, Zhang et al. combine VR--CARLA
simulation with eye tracking to study driver responses
\cite{zhang2025workzonesafety}, while VR-WISE integrates
CARLA--SUMO co-simulation with gaze and physiological
sensing to investigate driver awareness
\cite{zhang2025vrwise}.
These studies provide a methodological basis for
collecting human responses under controlled driving
conditions.

Work zones are not inherently safety-critical, but particular layouts,
environmental conditions, and worker interactions can increase conflict
potential. This motivates a trajectory-centered benchmark that
combines human-driving signals and the context of both work-zone under a matched
control condition and crash-informed, safety-critical configurations.

%%%%%%%%%%%%%%%%%%%%%%
\begin{figure*}[t]
    \centering
    \includegraphics[
        width=0.85\textwidth,
        trim=0bp 125bp 0bp 0bp,
        clip
    ]{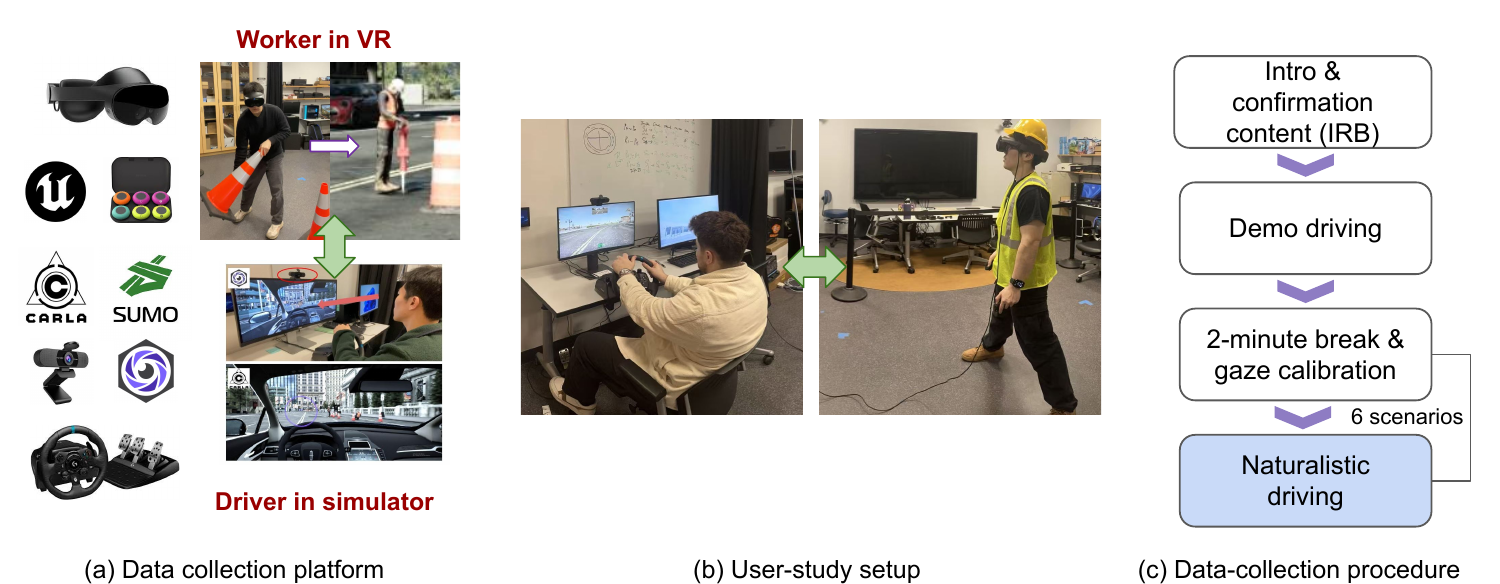}
    \caption{Overview of the \textsc{DualSense} platform used to collect
    \textsc{WZDrive-Data}. (a) Platform architecture connecting the human
    driver, VR roadway worker, simulation environment, and sensing devices.
    (b) User-study setup enabling real-time driver--worker interaction.
    (c) Data-collection procedure across six randomized scenario conditions.}
    \label{fig:data_collection}
\end{figure*}
%%%%%%%%%%%%%%%%%%%%%%%%%%%%%%%%%%%%%

%%%%%%%%%%%%%%%%%%%%%%%%%%%%%%%%%%%%%%%
\section{Dataset}
\label{sec:dataset}

\subsection{Human-in-the-Loop Data Collection }
\label{sec:data_collection}

\noindent\textbf{Data collection platform.}
We collected \textsc{WZDrive-Data} using a human-in-the-loop
platform integrating CARLA--SUMO co-simulation
\cite{dosovitskiy2017carla,lopez2018sumo},
physical driving controls, gaze tracking, scene sensing,
and a VR interface for roadway-worker interaction
(Fig.~\ref{fig:data_collection}).
The platform records synchronized human-driving and
scene data for ego-trajectory prediction in roadway
work zones. The collection platform is described on the
DuoRoad project website.\footnote{\url{https://shuo9898-zs.github.io/duoroad/}}

\noindent\textbf{Work-zone and scenario design.}
The study included three work-zone configurations.
WZ1 and WZ3 used prerecorded worker animations within
the work-zone boundary. In WZ2, a human VR operator
placed traffic cones to form a temporary boundary while
interacting with the driver in real time. Three students
with civil-engineering backgrounds served as VR-worker
operators, enabling dual-human-in-the-loop interaction. Each participant encountered six scenario conditions:
a standard lane closure (S1, control), absent warning
signage (S2), rain (S3), a curved road (S4), nighttime
driving (S5), and truck involvement (S6).
S2--S6 examine absent advance-warning signage, rain,
curved roads, nighttime driving, and truck involvement.
The weather, road geometry, lighting, and heavy-vehicle
conditions reflect factors examined in empirical
analyses of work-zone crashes
\cite{ghasemzadeh2019weather,li2009risk,islam2024geometry}. The baseline layout comprised
an 8~m coned work area with a warning sign 8~m upstream
of the first cone; S2 omitted the sign.
Each scenario used a distinct map with four to six
SUMO-controlled vehicles, all trucks in S6.
Their motion responded to surrounding traffic, including
the ego vehicle, through the Intelligent Driver Model (IDM).

% \noindent\textbf{Participants and study procedure.}
% We recruited 34 licensed drivers; one participant's data
% were excluded because of a sensor disconnection.
% The remaining 33 participants were 18--34 years old
% ($M=25.94$, $SD=3.42$), with a median driving experience
% of 5.0 years (IQR: 3.0--7.25).
% All participants provided informed consent and received
% a \$10 gift card. The study received institutional review
% board approval, with institutional details withheld
% for anonymous review.

\noindent\textbf{Participants and study procedure.}
We recruited 34 licensed drivers and excluded one
participant's data due to sensor disconnection.
The final sample comprised 33 drivers aged 18--34 years
($M=25.94$, $SD=3.42$), with a median driving experience
of 5.0 years ($Q_1=3.0$, $Q_3=7.25$).
The study was approved by our institutional review board
(IRB), whose identifying details are withheld for
anonymous review. All participants provided informed consent and received compensation. Participants used a Logitech G923 steering wheel and pedals
and were instructed to drive as they normally would.
The study procedure is summarized in
Fig.~\ref{fig:data_collection}(c).
Before recording, participants received instructions
and completed gaze calibration and a familiarization drive.
Experimental maps were previously unseen, and work-zone
locations were not disclosed. The six scenarios were
presented in randomized order, each lasting approximately
5 minutes and containing two encounters with each of WZ1--WZ3.
A 2-minute break separated scenarios to reduce fatigue
and carryover effects. After the experiment, participants
completed a demographic questionnaire.
Recordings were aligned by timestamp, and work-zone
encounter segments were extracted to construct the dataset.

%%%%%%%%%%%%%%%%%
\subsection{Dataset Characteristics}
\label{sec:dataset_characteristics}

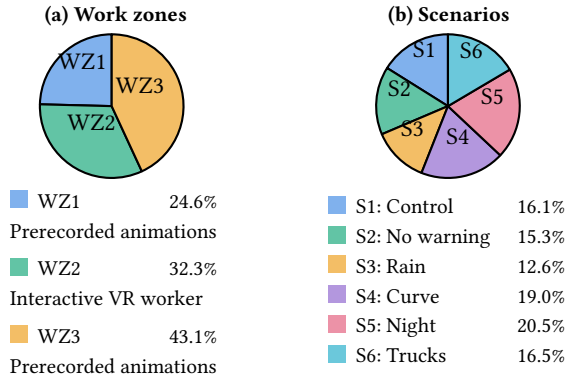
\begin{figure}[t]
\centering

\begin{minipage}[t]{0.48\linewidth}
\vspace{0pt}
\centering
{\small\textbf{(a) Work zones}\par}
\smallskip

\begin{tikzpicture}
\pie[
    radius=0.95,
    sum=100,
    rotate=90,
    hide number,
    text=inside,
    color={WZBlue,WZGreen,WZSand},
    every node/.style={font=\footnotesize}
]{
    24.6/WZ1,
    32.3/WZ2,
    43.1/WZ3
}
\end{tikzpicture}

\smallskip
{\small
\renewcommand{\arraystretch}{1.12}
\begin{tabular}{@{}lr@{}}
\wzlegend{WZBlue}{WZ1} & 24.6\% \\
\multicolumn{2}{@{}l@{}}{Prerecorded animations} \\[3pt]
\wzlegend{WZGreen}{WZ2} & 32.3\% \\
\multicolumn{2}{@{}l@{}}{Interactive VR worker} \\[3pt]
\wzlegend{WZSand}{WZ3} & 43.1\% \\
\multicolumn{2}{@{}l@{}}{Prerecorded animations} \\
\end{tabular}
}
\end{minipage}
\hfill
\begin{minipage}[t]{0.48\linewidth}
\vspace{0pt}
\centering
{\small\textbf{(b) Scenarios}\par}
\smallskip

\begin{tikzpicture}
\pie[
    radius=0.95,
    sum=100,
    rotate=90,
    hide number,
    text=inside,
    color={
        WZBlue,WZGreen,WZSand,
        WZPurple,WZRose,WZSlate
    },
    every node/.style={font=\footnotesize}
]{
    16.1/S1,
    15.3/S2,
    12.6/S3,
    19.0/S4,
    20.5/S5,
    16.5/S6
}
\end{tikzpicture}

\smallskip
{\small
\renewcommand{\arraystretch}{1.12}
\begin{tabular}{@{}lr@{}}
\wzlegend{WZBlue}{S1: Control}     & 16.1\% \\
\wzlegend{WZGreen}{S2: No warning} & 15.3\% \\
\wzlegend{WZSand}{S3: Rain}        & 12.6\% \\
\wzlegend{WZPurple}{S4: Curve}     & 19.0\% \\
\wzlegend{WZRose}{S5: Night}       & 20.5\% \\
\wzlegend{WZSlate}{S6: Trucks}     & 16.5\% \\
\end{tabular}
}
\end{minipage}

\caption{Distribution of prediction windows by
(a) work-zone configuration and (b) scenario condition.}
\label{fig:dataset_distribution}
\end{figure}

\noindent\textbf{Dataset scale and temporal coverage.}
\textsc{WZDrive-Data} comprises 1,183 distinct work-zone encounter
segments from 33 drivers, covering three work-zone configurations
and six scenario conditions. Each segment corresponds to one
participant, scenario, work zone, and lap. Together, these segments
provide approximately 3.7 hours of trajectory coverage and yield
26,892 overlapping prediction windows. As summarized in
Table~\ref{tab:dataset_summary}, each window contains 10 observed
frames and 25 future positions at 5~Hz, corresponding to a nominal
2~s observation history and a 5~s prediction horizon. Consecutive
windows are separated by 0.2~s. Figure~\ref{fig:dataset_distribution}
shows their distribution across work zones and scenarios.
Participant partitions are defined separately for each
experimental protocol.

\noindent\textbf{Multimodal content and representation.}
The structured inputs comprise six groups: ego motion, driver
controls, gaze, surrounding agents, lane geometry, and work-zone
context. Stored features contain 6 values per ego frame,
3 per control or gaze frame, and 11 per agent per frame,
with up to 20 agents. Lane geometry is represented by 8-D
point features, 10-D lane attributes, and graph connectivity.
Work-zone reference points and static worker anchors use
3-D features encoding relative position and validity.
Masks distinguish valid observations from missing or padded entries.
In addition to these structured records, the dataset retains
front-view RGB images, semantic-segmentation images, and
driving-view screenshots. Gaze and all three visual streams
are recorded at 10~Hz; gaze observations are aligned with
the 5~Hz trajectory sequence when constructing model inputs.

% \noindent\textbf{Relation to general-driving datasets.}
% For context, nuScenes contains 1,000 real-world scenes of
% 20~s each, totaling approximately 5.6 hours
% \cite{caesar2020nuscenes}. Its scenes and our work-zone encounter
% segments represent different collection units and should not be
% compared directly by count. Complementing broad real-world driving
% coverage, \textsc{WZDrive-Data} focuses on repeatable work-zone
% conditions with synchronized driver controls, gaze, trajectories,
% and explicit temporary scene geometry.

\noindent\textbf{Relation to general-driving datasets.}
For context, nuScenes contains 1,000 real-world scenes
of 20~s each, totaling approximately 5.6 hours
\cite{caesar2020nuscenes}.
\textsc{WZDrive-Data} adopts a nuScenes-style scene
and agent organization, with trajectories represented
as timestamped positions on the two-dimensional ground plane.
However, nuScenes scenes and our work-zone encounter
segments represent different collection units and
should not be compared directly by count.
Complementing broad real-world driving coverage,
\textsc{WZDrive-Data} focuses on repeatable work-zone
conditions with synchronized driver controls, gaze,
trajectories, and explicit temporary scene geometry.

\begin{table}[t]
\centering
\caption{\textsc{WZDrive-Data} scale and multimodality}
\label{tab:dataset_summary}
\small

\textbf{(a) Dataset scale}\par
\smallskip
\begin{tabular*}{\linewidth}{
    @{\extracolsep{\fill}}lr@{}
}
\toprule
Characteristic & Value \\
\midrule
Valid drivers & 33 \\
Work-zone configurations & 3 \\
Scenario conditions & 6 \\
Distinct encounter segments & 1,183 \\
Overlapping prediction windows & 26,892 \\
Trajectory coverage & $\approx 3.7$~h \\
Trajectory sampling rate & 5~Hz \\
Observation / prediction horizon & 2~s / 5~s \\
\bottomrule
\end{tabular*}

\medskip
\textbf{(b) Multimodal content}\par
\smallskip
\begin{tabular*}{\linewidth}{
    @{\extracolsep{\fill}}lr@{}
}
\toprule
Modality / scene information & Rate / representation \\
\midrule
Ego-vehicle motion & 5~Hz \\
Driver controls & 5~Hz \\
Driver gaze & 10~Hz \\
Surrounding-agent trajectories & 5~Hz \\
Interactive VR-worker pose & 5~Hz \\
Front-view RGB images & 10~Hz \\
Semantic-segmentation images & 10~Hz \\
Driving-view screenshots & 10~Hz \\
Lane geometry and connectivity & Vector map \\
Work-zone geometry and worker animations & Scene context \\
\bottomrule
\end{tabular*}

\smallskip
\begin{minipage}{\linewidth}
\footnotesize
Trajectory coverage counts overlapping windows only once
within each encounter segment. Gaze is recorded at 10~Hz
and aligned to 5~Hz for model input. Static geometry is
represented spatially rather than assigned a sampling rate.
\end{minipage}
\end{table}

%%%%%%%%%%%%%%%%%%%%%%%%%%
%%%%%%%%%%%%%%%%%%%%%%%%%%%%%%%%%%%%%%%%

\section{Methodology}
\label{sec:method}
\label{sec:wzdrive_model}

\begin{figure*}[t]
    \centering
    \includegraphics[
        width=0.78\textwidth,
        trim={0 0 3.5cm 0},
        clip
    ]{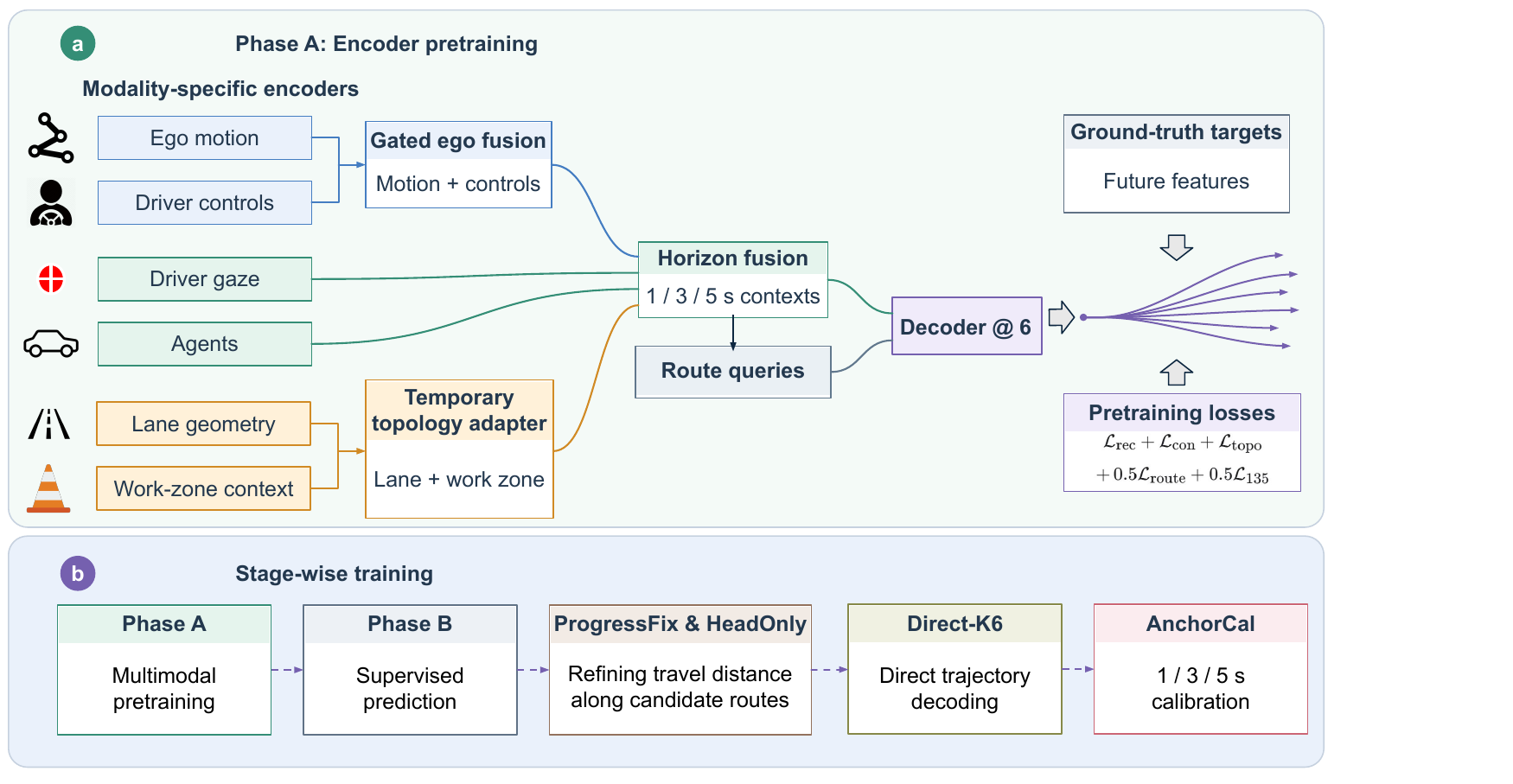}
    \caption{Overview of \textsc{WZDrive} pretraining and staged training.
    (a) Modality-specific encoders and fusion modules learn representations
    through reconstruction, contrastive, topology, route-coverage, and
    travel-distance objectives.
    (b) The six-stage training schedule, from encoder pretraining to
    temporal anchor calibration.}
    \label{fig:wzdrive_pipeline}
\end{figure*}

We propose \textsc{WZDrive}, a multimodal ego-trajectory prediction model
that combines driver behavior with roadway-work-zone context. The model
encodes vehicle motion, driver-control history, gaze, and scene geometry
to generate six candidate future trajectories and their probabilities.
A temporal anchor-calibration module subsequently adjusts the predicted
positions to address residual prediction errors. Figure~\ref{fig:wzdrive_pipeline} summarizes the encoder pretraining
framework and subsequent training stages.

%%%%%%%%%%%%%%%%%%%%%%%%%%%
% \begin{figure*}[t]
%     \centering
%     \includegraphics[width=0.8\textwidth]{Figures/WZDrive_pipeline.pdf}
%     \caption{Overview of \textsc{WZDrive} pretraining and staged training.
%     (a) Modality-specific encoders and fusion modules learn representations
%     through reconstruction, contrastive, topology, route-coverage, and
%     travel-distance objectives.
%     (b) The six-stage training schedule, from encoder pretraining to
%     temporal anchor calibration. Arrows between stages indicate
%     parameter transfer rather than sequential inference.}
%     \label{fig:wzdrive_pipeline}
% \end{figure*}

\subsection{Problem Formulation}
\label{sec:prediction_task}

Given 2~s of observations, we predict the human-driven ego
vehicle's trajectory over the next 5~s. At 5~Hz, these
windows contain $T_{\mathrm{obs}}=10$ observed frames and
$T_{\mathrm{pred}}=25$ future frames. The input is
\begin{equation}
\mathcal{X} =
\left\{
\mathbf{X}^{\mathrm{ego}},
\mathbf{X}^{\mathrm{ctrl}},
\mathbf{X}^{\mathrm{gaze}},
\mathbf{X}^{\mathrm{agents}},
\mathcal{L}, \mathcal{W}
\right\},
\end{equation}
where $\mathbf{X}^{\mathrm{ego}}$ denotes ego-motion history;
$\mathbf{X}^{\mathrm{ctrl}}$ contains historical steering,
throttle, and brake signals; and
$\mathbf{X}^{\mathrm{gaze}}$ denotes driver-gaze history.
$\mathbf{X}^{\mathrm{agents}}$ contains the motion histories
of surrounding vehicles and VR-controlled workers.
$\mathcal{L}$ represents lane geometry, while $\mathcal{W}$
contains work-zone geometry and static worker context.
All temporal inputs are restricted to the observation window.
Trajectory coordinates are expressed in a local frame
centered at the ego vehicle's latest observed position
and aligned with its heading. To represent alternative future motions, \textsc{WZDrive}
outputs $K=6$ candidate trajectories:
\begin{equation}
\hat{\mathcal{Y}} =
\left\{
\hat{\mathbf{Y}}^{(1)}, \ldots, \hat{\mathbf{Y}}^{(K)}
\right\},
\qquad
\hat{\mathbf{Y}}^{(k)}
\in \mathbb{R}^{T_{\mathrm{pred}}\times 2}.
\end{equation}
Each candidate contains 25 two-dimensional positions
at 0.2~s intervals following the latest observation.

%%%%%%%%%%%%%%
\subsection{Multimodal Encoding and Fusion}
\label{sec:multimodal_encoding}

Separate encoders process ego-motion history,
driver-control history, gaze history, and surrounding-agent
histories, producing $\mathbf{z}_{\mathrm{motion}}$,
$\mathbf{z}_{\mathrm{control}}$,
$\mathbf{z}_{\mathrm{gaze}}$, and
$\mathbf{z}_{\mathrm{agents}}$, respectively.
Lane geometry and work-zone context are encoded as
$\mathbf{z}_{\mathrm{lane}}$ and
$\mathbf{z}_{\mathrm{WZ}}$.
The agent branch includes surrounding vehicles and
VR-controlled workers, while the work-zone branch
includes temporary geometry and static worker information.

Motion and control features are combined through learned
scoring to form $\mathbf{z}_{\mathrm{motion+control}}$.
The scores weight control features to incorporate
relevant driving actions into the motion representation,
excluding invalid control observations.
The model then uses $\mathbf{z}_{\mathrm{motion+control}}$
to compute a relevance score for each agent and aggregate
their features accordingly. Individual agent features
are also retained so that the decoder can attend to
specific vehicles and workers. Gaze is encoded separately
and participates in the subsequent fusion.

Lane and work-zone representations are combined through
a temporary topology adapter. Using lane features and
$\mathbf{z}_{\mathrm{WZ}}$, the adapter learns two types
of scores: one assesses the suitability of each lane
for travel under current work-zone conditions, while
the other assesses the suitability of connections
between lanes. Connection scores additionally consider
the relative position, orientation, and connection type
of the two lanes. These scores act as soft weights
that jointly regulate information exchange within the
lane graph. The updated lane features are pooled into
$\mathbf{z}_{\mathrm{map}}$, representing the road layout
as affected by the work zone. Finally, horizon-aware fusion combines the four
representations:
\begin{equation}
\mathbf{c}_h = \mathcal{F}_h\bigl(
\mathbf{z}_{\mathrm{motion+control}},
\mathbf{z}_{\mathrm{gaze}},
\mathbf{z}_{\mathrm{agents}},
\mathbf{z}_{\mathrm{map}}
\bigr).
\end{equation}
Here, $h\in\{1,3,5\}$ denotes the prediction horizon
in seconds.
The contexts $\mathbf{c}_1$, $\mathbf{c}_3$, and
$\mathbf{c}_5$ guide predictions over future intervals
$(0,1]$, $(1,3]$, and $(3,5]$~s, respectively.
Each context uses horizon-dependent modality weights,
allowing different intervals to draw on different
information combinations. The decoder predicts the
complete 5~s trajectory at 0.2~s intervals, yielding
25 positions rather than only three endpoints.
%%%%%%%%%%%%%%

\noindent\textbf{Trajectory decoder.}
Inspired by MTR's use of mode-specific queries
\cite{shi2022mtr}, the decoder generates six candidate
trajectories over 5~s at 5~Hz, each containing 25
ego-local positions at 0.2~s intervals.
A learned motion-prior network uses
$\mathbf{z}_{\mathrm{motion+control}}$ and
$\mathbf{z}_{\mathrm{control}}$ to estimate all 25 future
velocities. Multiplying these by $\Delta t=0.2$~s yields
prior displacements $\Delta\mathbf{y}^{\mathrm{prior}}_t$
shared across candidates.

Each candidate has 25 time-indexed queries combining
motion--control features, a candidate embedding,
time encoding, the corresponding fused context
$\mathbf{c}_h$, and encoded prior position and velocity.
Cross-attention updates these queries using ego, lane,
agent, worker, and work-zone features, with invalid
features masked. All candidates and future steps are
processed in a single forward pass, rather than
autoregressively, to predict candidate-specific
displacement residuals $\mathbf{r}_{k,t}$.
Adding these residuals to the shared prior displacements
and accumulating in temporal order gives the initial
predicted positions:
\begin{equation}
\hat{\mathbf{y}}^{\mathrm{init}}_{k,t}
=\sum_{\tau=1}^{t}
(\Delta\mathbf{y}^{\mathrm{prior}}_{\tau}
+\mathbf{r}_{k,\tau}).
\end{equation}
Here, $k\in\{1,\ldots,6\}$ indexes candidates and
$t\in\{1,\ldots,25\}$ indexes future steps.
A scoring head assigns candidate probabilities,
and temporal anchor calibration further adjusts
the initial trajectories
(Section~\ref{sec:training_refinement}).

\subsection{Stage-wise Training and Trajectory Refinement}
\label{sec:training_refinement}

Training proceeds from multimodal pretraining to supervised
prediction and subsequent refinement. Early stages use
candidate routes through the lane graph, with route progress
specifying how far the vehicle travels along each route.
Later stages introduce the direct trajectory decoder and
calibrate its predictions.

\noindent\textbf{Phase A: Multimodal pretraining.}
Inspired by masked pretraining
\cite{cheng2023forecastmae} and contrastive learning
for trajectory prediction \cite{yao2024trajclip},
we pretrain the encoders, fusion modules, and
route-prediction components using
\begin{equation}
\mathcal{L}_{\mathrm{PhaseA}}
=
\mathcal{L}_{\mathrm{rec}}
+\mathcal{L}_{\mathrm{con}}
+\mathcal{L}_{\mathrm{topo}}
+0.5\mathcal{L}_{\mathrm{route}}
+0.5\mathcal{L}_{135}.
\end{equation}
The reconstruction loss $\mathcal{L}_{\mathrm{rec}}$
teaches the model to recover masked input information.
The contrastive loss $\mathcal{L}_{\mathrm{con}}$
helps the model associate observed driving context
with its corresponding future behavior.
The topology loss $\mathcal{L}_{\mathrm{topo}}$
supervises lane connections and the spatial relationships
between lanes and work-zone elements.
The route-coverage loss $\mathcal{L}_{\mathrm{route}}$
encourages candidate routes to include a path close
to the driver's actual route.
The travel-distance loss $\mathcal{L}_{135}$
supervises the cumulative distance the vehicle
travels by 1, 3, and 5~s.

\noindent\textbf{Phase B: Supervised trajectory prediction.}
Whereas Phase A learns representations and route
information, Phase B trains complete trajectory
predictions through direct position supervision:
\begin{equation}
\mathcal{L}_{\mathrm{PhaseB}}
=
\mathcal{L}_{\mathrm{traj}}
+0.5\mathcal{L}_{\mathrm{end}}
+\mathcal{L}_{\mathrm{aux}}
+0.2\mathcal{L}_{\mathrm{local\text{-}refine}}.
\end{equation}
The trajectory loss $\mathcal{L}_{\mathrm{traj}}$
penalizes position errors at all 25 future steps,
while the endpoint loss $\mathcal{L}_{\mathrm{end}}$
emphasizes the final position.
Both additionally penalize longitudinal errors.
The auxiliary loss $\mathcal{L}_{\mathrm{aux}}$
combines weighted objectives for route prediction,
candidate ranking, and consistency with lanes,
work zones, and surrounding agents.
The refinement loss
$\mathcal{L}_{\mathrm{local\text{-}refine}}$
trains the refiner to correct coarse trajectory
position errors.
The trajectory and endpoint losses are reused
in subsequent training stages.

\noindent\textbf{Route-progress refinement.}
Inspired by PGP's route-conditioned prediction
\cite{deo2022pgp}, we refine route progress in two stages. \textbf{ProgressFix} refines the predicted distance
traveled along a candidate route by 1, 3, and 5~s
using $\mathcal{L}_{\mathrm{progress}}$.
\textbf{HeadOnly} further refines these predictions
by learning corrections to the travel increment
at each of the 25 future steps.

\noindent\textbf{Direct-K6: Direct trajectory prediction.}
Direct-K6 refines six-candidate trajectory prediction
to improve accuracy and promote diversity among
candidate routes.
Starting from the HeadOnly checkpoint, we introduce
a new direct decoder and jointly train it with
the existing modules.
The training objective combines trajectory loss
$\mathcal{L}_{\mathrm{traj}}$, endpoint loss
$\mathcal{L}_{\mathrm{end}}$, candidate-scoring loss
$\mathcal{L}_{\mathrm{score}}$, motion-prior loss
$\mathcal{L}_{\mathrm{prior}}$, and route-diversity loss
$\mathcal{L}_{\mathrm{div}}$.

\noindent\textbf{AnchorCal: Temporal position calibration.}
Similar to the residual trajectory refinement in
LAFormer \cite{liu2024laformer}, AnchorCal learns
position corrections to improve the initial predictions.
We freeze the Direct-K6 predictor and attach a
calibration head to its decoder features.
The head predicts longitudinal and lateral position
corrections at 1, 3, and 5~s.
These corrections are linearly interpolated to all
25 future steps and added to the original predictions:
\begin{equation}
\hat{\mathbf{y}}^{\mathrm{final}}_{k,t}
=
\hat{\mathbf{y}}^{\mathrm{init}}_{k,t}
+\boldsymbol{\delta}_{k,t}.
\end{equation}
Here, $\boldsymbol{\delta}_{k,t}$ denotes the interpolated
position correction, with zero correction at the current
time.
The calibrated trajectories are compared with the
ground truth using $\mathcal{L}_{\mathrm{traj}}$ and
$\mathcal{L}_{\mathrm{end}}$, updating only the
calibration head.

%%%%%%%%%%%%%%%%%%%%%%%%%%%%%%%%%%%%%
\section{Results}
\label{sec:results}

%%%%%%%%%%%%%%%%%%%%%%%%%%%
\begin{table}[t]
\centering
\caption{Rule-based and time-series baseline performance
($K=1$) with selected input combinations. Lower is better.}
\label{tab:sequence_baselines}
\small
\setlength{\tabcolsep}{3pt}
\begin{tabular}{@{}llrr@{}}
\toprule
Model & Features & ADE (m) $\downarrow$ & FDE (m) $\downarrow$ \\
\midrule
Stay & Ego & 21.635 & 42.053 \\
CV & Ego & 7.237 & 16.896 \\
\midrule
LSTM & Ego + control + WZ & 2.988 & 7.513 \\
Transformer & Ego + control + WZ & 2.876 & \textbf{7.284} \\
GRU & Ego + control + WZ + lane & \textbf{2.860} & 7.320 \\
\bottomrule
\end{tabular}
\end{table}

%%%%%%%%%%%%%%%%%%%%%%%%%%%%%%%%%%%%%
\begin{table*}[t]
\centering
\caption{Comparison with SOTA baselines ($K=6$).
Errors are in meters.}
\label{tab:multimodal_baselines}
\normalsize
\setlength{\tabcolsep}{6pt}
\renewcommand{\arraystretch}{1.18}

\begin{tabular}{@{}lrrrr@{}}
\toprule
& \multicolumn{2}{c}{Without WZ}
& \multicolumn{2}{c}{With WZ} \\
\cmidrule(lr){2-3}
\cmidrule(lr){4-5}
Model
& minADE$_6$ $\downarrow$
& minFDE$_6$ $\downarrow$
& minADE$_6$ $\downarrow$
& minFDE$_6$ $\downarrow$ \\
\midrule

PGP (2021) \cite{deo2022pgp}
& 1.286 & 2.749
& 1.260 ($\downarrow$2.02\%)
& 2.742 ($\downarrow$0.25\%) \\

MTR (2022) \cite{shi2022mtr}
& 1.317 & 2.660
& 1.299 ($\downarrow$1.37\%)
& 2.567 ($\downarrow$3.50\%) \\

Forecast-MAE (2023) \cite{cheng2023forecastmae}
& 1.373 & 2.650
& 1.264 ($\downarrow$7.94\%)
& 2.475 ($\downarrow$6.60\%) \\

% QCNet (2023) \cite{zhou2023qcnet}
% & -- & --
% & 1.465 & 2.706 \\

LAFormer (2024) \cite{liu2024laformer}
& \underline{1.204} & 2.662
& \underline{1.179} ($\downarrow$2.08\%)
& 2.550 ($\downarrow$4.21\%) \\

DeMo (2024) \cite{zhang2024demo}
& 1.514 & \underline{2.436}
& 1.281 ($\downarrow$15.39\%)
& \underline{2.462} ($\uparrow$1.07\%) \\

RouteFormer (2025) \cite{akbiyik2025routeformer}
& 2.106 & 4.023
& 2.050 ($\downarrow$2.65\%)
& 3.847 ($\downarrow$4.38\%) \\

\midrule
MM-GRU (ours)
& 1.234 & 2.682
& 1.124 ($\downarrow$8.91\%)
& 2.242 ($\downarrow$16.41\%) \\

\midrule
\textsc{WZDrive} (ours)$^{\dagger}$
& -- & --
& $\mathbf{1.094 \pm 0.022}$ ($\downarrow$7.18\%)
& $\mathbf{2.115 \pm 0.082}$ ($\downarrow$13.18\%) \\

\bottomrule
\end{tabular}

\smallskip
\begin{minipage}{0.95\textwidth}
\small
Baseline percentages compare with the same model without WZ.
$^{\dagger}$WZDrive percentages compare with the best published
baselines: LAFormer with WZ for minADE$_6$ and DeMo without WZ
for minFDE$_6$.
WZDrive reports mean $\pm$ SD over three seeds.
Underlining marks the best published baseline per column,
excluding MM-GRU; bold marks the best overall means.
\end{minipage}
\end{table*}
%%%%%%%%%%%%%%%%%%%%%%%%%%%%%%%%%%%%%

\subsection{Experimental Setup}
\label{sec:experimental_setup}

\noindent\textbf{Protocol and implementation.}
All models observe 2~s and predict 5~s of ego motion at 5~Hz.
The main comparison uses participant-disjoint Set1:
27 training drivers, P14/P20/P34 for validation, and
P2/P12/P31 for testing, yielding 22,540/2,119/2,233
windows, respectively.
Experiments used an Intel Core i7-14700K CPU,
64~GB RAM, and an NVIDIA RTX 4090 GPU.

\noindent\textbf{Baselines and metrics.}
We compare rule-based Stay/CV, GRU, LSTM, Transformer,
and our MM-GRU baseline. Advanced methods represent
visual and gaze modeling (RouteFormer),
lane-aware prediction (LAFormer, PGP), query-based
prediction (MTR), intention--dynamics decomposition
(DeMo), and masked pretraining (Forecast-MAE);
citations are provided in Table~\ref{tab:multimodal_baselines}.
We evaluate WZ and no-WZ variants where available.
Single-trajectory baselines use ADE/FDE; six-candidate
models use minADE$_6$/minFDE$_6$, which independently
select the lowest average trajectory and endpoint errors.
Results are averaged over test samples in meters;
lower is better.

%%%%%%%%%%%%%%%
\subsection{Trajectory Prediction Performance}
\label{sec:prediction_results}

\noindent\textbf{Rule-based and time-series baselines.}
Table~\ref{tab:sequence_baselines} compares single-trajectory
predictors ($K=1$), including stationary (Stay),
constant-velocity (CV), and time-series models.
For each learned model, we report its best-performing
input combination among those evaluated.
All learned models outperform Stay and CV.
GRU achieves the lowest ADE, while Transformer achieves
the lowest FDE, reflecting both architecture and input differences.

\noindent\textbf{Multimodal forecasting baselines.}
Table~\ref{tab:multimodal_baselines} compares methods
standardized to six candidate trajectories ($K=6$).
We evaluate variants with and without work-zone information,
incorporating it through an additional encoder or the existing
lane encoder and reporting the best-performing integration
evaluated for each model.
Work-zone information reduces minADE for every paired baseline
and minFDE for all except DeMo, demonstrating its value
for prediction in this setting.

Among published baselines, LAFormer achieves the lowest
minADE, consistent with its lane-aware design, while DeMo
achieves the lowest minFDE.
RouteFormer has the highest errors on both metrics.
Its reliance on RGB imagery and gaze may make learning
more demanding under our limited-data setting.
Motivated by GRU's competitive time-series performance,
we develop MM-GRU to predict six candidate trajectories
and evaluate different input combinations.
The selected configuration combines ego motion, driver controls,
surrounding agents, lane geometry, and work-zone context,
including worker information.
MM-GRU outperforms the published baselines, supporting
the utility of multimodal inputs.
Together, these results suggest that effective representation
and fusion matter more than simply adding modalities.
MM-GRU outperforms the published baselines, supporting
the utility of multimodal inputs.
Together, these results suggest that effective representation
and fusion matter more than simply adding modalities.

WZDrive achieves the lowest mean errors across three seeds:
$1.094 \pm 0.022$~m minADE$_6$ and
$2.115 \pm 0.082$~m minFDE$_6$.
Relative to the best published baseline for each metric,
it reduces minADE by 7.18\% compared with LAFormer with WZ
and minFDE by 13.18\% compared with DeMo without WZ.

%%%%%%%%%%%%%%%%%%%%%%%%%%

\subsection{Ablation Studies}
\label{sec:ablation}

\noindent\textbf{Input-modality ablation.}
We mask the selected inputs and retrain each variant
from scratch using the same training schedule.
The WZ variant also disables the associated topology
adaptation and auxiliary losses.
Table~\ref{tab:ablation}(a) shows that removing gaze
has little effect in this setting.
Removing driver controls causes the largest degradation,
increasing minADE and minFDE by 17.20\% and 14.17\%,
respectively. Removing WZ and worker context and its
associated mechanisms increases these errors by
7.54\% and 12.78\%, supporting the value of human
control signals and work-zone-aware modeling.

\noindent\textbf{Stage-wise training.}
Table~\ref{tab:ablation}(b) shows the effectiveness
of the overall training pipeline.
Although intermediate stages do not improve monotonically,
Direct-K6 produces the largest error reduction,
and AnchorCal further improves most metrics.
The final model outperforms Phase B across all three
participant partitions.

\noindent\textbf{Participant generalization.}
We additionally evaluate two participant-disjoint partitions:
Set2 uses P5/P15/P18 for validation and P3/P4/P25 for testing;
Set3 uses P7/P11/P19 and P13/P17/P23, respectively.
The remaining 27 participants form each training partition.
Both show an overall error-reduction trend similar to Set1,
while differences in absolute performance may reflect
variation in driving behavior across participant groups.

%%%%%%%%%%%%%%%%%%%%%%%%%%
\begin{table}[]
\centering
\caption{Effects of input modalities and training stages
across participant partitions. Errors are in meters;
lower is better.}
\label{tab:ablation}
\normalsize
\renewcommand{\arraystretch}{1.15}

\textbf{(a) Input modalities on Set1}\par
\label{tab:modality_ablation}
\smallskip
\setlength{\tabcolsep}{4pt}
\begin{tabular}{@{}lrr@{}}
\toprule
Variant & minADE$_6$ & minFDE$_6$ \\
\midrule
\textsc{WZDrive} (full)
& \textbf{1.103} & 2.119 \\
Without gaze
& 1.108 & \textbf{2.116} \\
Without driver controls
& 1.293 & 2.419 \\
Without WZ and workers
& 1.187 & 2.390 \\
\bottomrule
\end{tabular}

\medskip
\textbf{(b) Training stages across participant partitions}\par
\label{tab:stagewise_ablation}
\smallskip
\setlength{\tabcolsep}{3pt}
\begin{tabular}{@{}lccc@{}}
\toprule
Stage & Set1 & Set2 & Set3 \\
\midrule
Phase B
& 1.572 / 3.692
& 1.900 / 4.156
& 1.989 / 4.133 \\
ProgressFix
& 1.621 / 3.827
& 1.910 / 4.207
& 1.981 / 4.377 \\
HeadOnly
& 1.591 / 3.614
& 1.913 / 4.169
& 2.001 / 4.341 \\
Direct-K6
& 1.148 / 2.244
& 1.418 / 2.749
& 1.530 / \textbf{2.801} \\
AnchorCal
& \textbf{1.094} / \textbf{2.115}
& \textbf{1.345} / \textbf{2.539}
& \textbf{1.496} / 2.809 \\
\bottomrule
\end{tabular}

\smallskip
\begin{minipage}{\linewidth}
\small
(a) Each input variant is retrained from scratch.
(b) Entries report minADE$_6$ / minFDE$_6$ across
three participant partitions.
AnchorCal is the final \textsc{WZDrive} model.
\end{minipage}
\end{table}
%%%%%%%%%%%%%%%%%%%%%%%%%%

\section{Discussion}
\label{sec:discussion}
%%%%%%%%%%%%%%%%

\begin{figure*}[t]
    \centering
    \includegraphics[width=\textwidth]{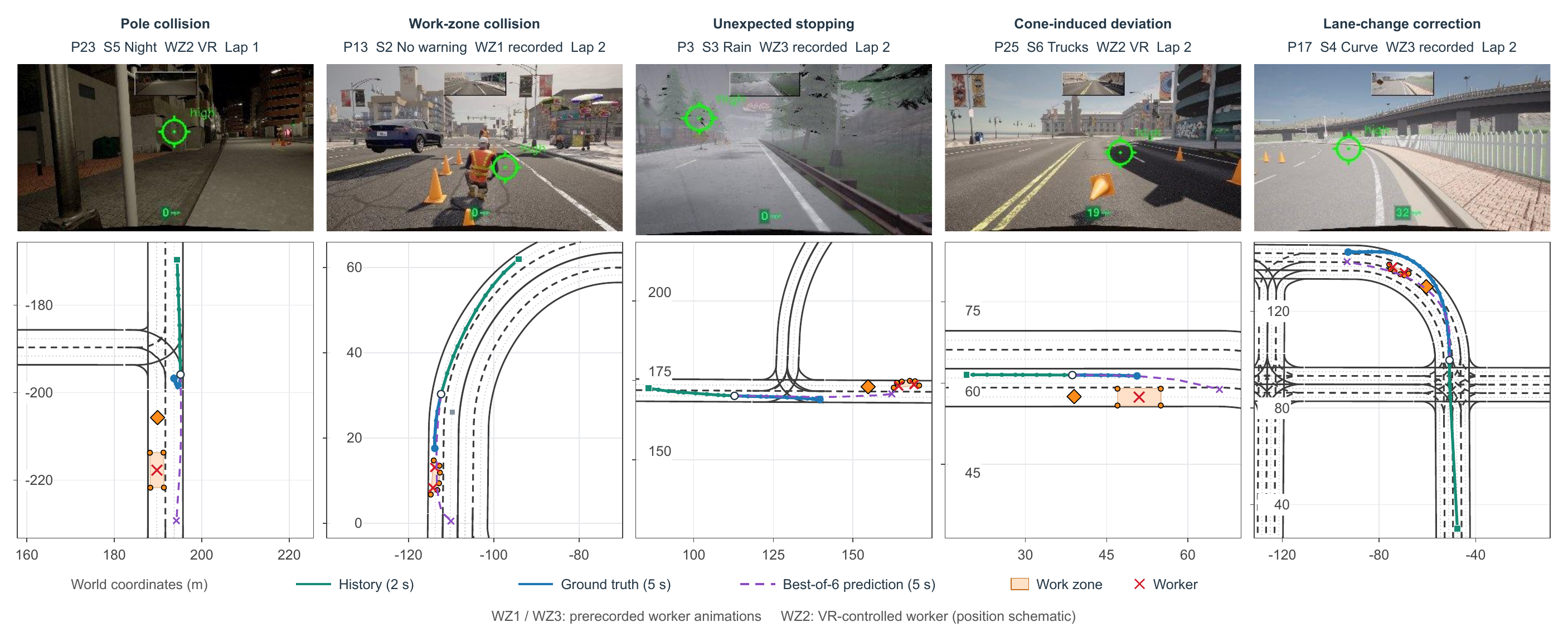}
    \caption{Qualitative failure cases. From left to right:
    pole collision, work-zone collision, unexpected stopping,
    cone-induced deviation, and lane-change correction.
    The top row shows driving views with gaze overlays;
    the bottom row shows observed motion, ground-truth futures,
    and the lowest-ADE prediction among six candidates.
    VR-worker markers indicate schematic positions.}
    \label{fig:failure_cases}
\end{figure*}
%%%%%%%%%%%%%%%%%%

\noindent\textbf{Gaze and driver behavior.}
Although gaze provides limited additional improvement in our
ablation, this does not mean that driver attention is unimportant.
The five safety-critical scenarios involve challenges arising
from scenario design, temporary geometry, and VR-worker
interactions, rather than ordinary driving conditions alone.
In these settings, visual attention is relevant to how drivers
perceive their surroundings, make decisions, and adjust their
driving behavior. Previous work-zone studies have similarly
examined the relationship between gaze, driver awareness,
and worker behavior
\cite{zhang2025workzonesafety,zhang2025vrwise}.
We also explored a model combining semantic images with gaze
instead of RGB images, but its performance remained below
\textsc{WZDrive}.
Together, these observations suggest that the behavioral value
of gaze is not automatically translated into prediction gains:
effectively incorporating attention and visual context remains
important for multimodal trajectory prediction.

% \noindent\textbf{The contribution of refinement.}
% The stage-wise results highlight the value of refinement.
% Route-progress refinement adjusts travel along candidate
% routes, while Direct-K6 introduces direct prediction of
% six trajectories using a shared motion prior.
% AnchorCal then adds learned position corrections at
% 1, 3, and 5~s, interpolated across all future steps.
% Although improvements are not monotonic, Direct-K6
% produces the largest error reduction, and AnchorCal
% further improves most metrics. Similar overall trends across participant partitions support the effectiveness of the training pipeline.

\noindent\textbf{Stage-wise training and trajectory refinement.}The stage-wise results in Table~\ref{tab:modality_ablation}(b)
highlight the value of the training and trajectory
refinement described in Section~4.3.
Route-progress refinement adjusts travel along candidate
routes, while Direct-K6 introduces direct prediction of
six trajectories using a shared motion prior.
AnchorCal then adds learned position corrections at
1, 3, and 5~s, interpolated across all future steps.
Although improvements are not monotonic, Direct-K6
produces the largest error reduction.
Relative to Direct-K6, AnchorCal further reduces
minADE$_6$ across all three partitions and minFDE$_6$
in Set1 and Set2, while minFDE$_6$ in Set3 increases slightly. Similar overall trends across participant partitions support the effectiveness of the training pipeline.

\noindent\textbf{Representative failure cases.}
Figure~\ref{fig:failure_cases} shows five cases selected
from ten high-error candidates across the test partitions
of Set1--Set3. Each represents a distinct encounter,
excluding repeated sliding windows from the same trajectory. The top row shows driving views with gaze overlays;
green markers indicate high-confidence gaze measurements.
The bottom row shows observed trajectories, ground-truth
futures, and predictions.

Based on video inspection, the first case shows a collision
near the work zone with an unclear cause.
In the second, the driver attempts to overtake a
SUMO-controlled vehicle blocking the view of the work zone.
Missing advance-warning signage in S2 further limits
early warning before a work-zone collision.
In the third, the driver stops to observe the work zone
without a collision or interference from another vehicle,
possibly reflecting caution in rain.
In the fourth, a cone enters the roadway during a Lap~2
VR-worker interaction, triggering abrupt braking.
Some hazardous interactions were intentionally introduced
in Lap~2 to enrich the dataset.
Finally, in the fifth case, the curve and nearby work zone
make a lane-change maneuver difficult, causing the vehicle
to drive onto the curb.

These cases highlight the difficulty of predicting abrupt
or unusual driving responses near work zones.
Although the scenarios were designed, the specific collisions,
stops, and corrective actions arose from human driving
rather than being prescribed.
These observations motivate broader training coverage
of such responses and highlight the value of simulation
for collecting them safely.

\noindent\textbf{Applications and pathway to deployment.}
WZDrive could support ADAS in roadway work zones by
informing warnings and corrective interventions when
drivers make control errors. Its predictions could also
inform downstream planning and control modules in
autonomous driving.
For Level 2--3 applications, the control history records
steering, throttle, and braking inputs over the preceding
2~s, originating from either the driver or the driving
system depending on who controls the vehicle.
During Level 4--5 autonomous operation, these inputs
are generated by the vehicle controller rather than
a human driver.
For on-vehicle deployment, historical control signals
can be obtained through the Controller Area Network
(CAN) bus and decoded into numerical features.
Vehicle-specific pre-processing aligns their definitions,
units, ranges, and sampling rates with the model inputs.
The gaze would be supplied separately by an eye-tracking
system and could be optional depending on the driver.

%%%%%%%%%%%%%%%%%%%%%%%%%%%%5
\section{Conclusion and Future Work}
\label{sec:conclusion}

This study addresses the prediction of the ego-trajectory in urban
roadway work zones. Unlike regular driving environments,
temporary road work and associated worker activities change how
vehicles navigate and interact with their surroundings.
Studying prediction under these conditions requires data
that capture both work-zone context and human driving
responses. We introduce \textsc{WZDrive-Data}, collected
through human-in-the-loop driving and VR-worker interaction,
establish a trajectory-prediction benchmark with adapted
baselines, and develop \textsc{WZDrive}, a multimodal
predictor with stage-wise training and refinement.
The results demonstrate the value of driver-control
signals and explicit work-zone context for this task.

Future work will address two main directions.
First, the current data were collected in a simulation tool developed by the authors os this paper 
from 33 drivers aged 18--34. Broader recruitment across
age groups and driving experience, together with
real-world data, is needed to assess generalization.
Second, although the data set covers three work-zone
configurations and six scenarios, rare events such as
collisions, atypical stopping, and unexpected worker
interactions remain sparsely represented.
Expanding work-zone layouts and interaction conditions
could provide additional examples of these events
for more comprehensive training and evaluation.

Overall, the simulated data set (\textsc{WZDrive-Data}), the work-zone ego-trajectory prediction benchmark, and the model (\textsc{WZDrive}) provide a foundation for studying ego-vehicle trajectory prediction in
urban work zones. The human-interaction-based data collection
approach could also be extended to settings such as
vehicle-pedestrian encounters, supporting the study
of driving responses in complex environments and
the development of autonomous vehicles that can better navigate work zones.

%%%%%%%%%%%%%%%%%%%%%%%%%%%%%%%
\newpage

\bibliographystyle{unsrt}
\bibliography{reference}

\end{document}